# OmniNeRF: Hybriding Omnidirectional Distance and Radiance fields for Neural Surface Reconstruction

Jiaming Shen[a,†], Bolin Song[b,†], Zirui Wu[*c,d,†], Yi Xu[e,†]

[a]The University of Sydney, Sydney, Australia; [b] Massachusetts Institute of Technology, Cambridge, MA, United States; [c]Institute for AI Industry Research, Tsinghua University, Beijing, China; [d]Beijing Institute of Technology, Beijing, China; [e]The University of New South Wales, Sydney, Australia

[*] Corresponding author: wuzirui@bit.edu.cn
[†]These authors contributed equally.

## ABSTRACT

3D reconstruction from images has wide applications in Virtual Reality and Automatic Driving, where the precision requirement is very high. Ground-breaking research in the neural radiance field (NeRF) by utilizing Multi-Layer Perceptions has dramatically improved the representation quality of 3D objects. Some later studies improved NeRF by building truncated signed distance fields (TSDFs) but still suffer from the problem of blurred surfaces in 3D reconstruction. In this work, this surface ambiguity is addressed by proposing a novel way of 3D shape representation, OmniNeRF. It is based on training a hybrid implicit field of Omni-directional Distance Field (ODF) and neural radiance field, replacing the apparent density in NeRF with omnidirectional information. Moreover, we introduce additional supervision on the depth map to further improve reconstruction quality. The proposed method has been proven to effectively deal with NeRF defects at the edges of the surface reconstruction, providing higher quality 3D scene reconstruction results.
**Keywords:** 3D Reconstruction, neural implicit representation, omnidirectional distance fields, neural radiance fields

## INTRODUCTION

Reconstructing high-quality 3D surfaces from a series of posed RGBD images is a fundamental problem in 3-dimensional vision and graphics. Recently, research on neural networks to represent 3D surfaces as neural implicit fields (radiance fields[1] or signed distance fields[2]) has made impressive progress. Traditional methods of expressing 3D scenes by using neural radiance fields (NeRF)1 usually use RGB images captured by cameras at different positions and viewing directions as well as camera poses generated from the pictures to train neural networks. The generated networks are often used to synthesize new views.

The NeRF function is a neural network that represents a continuous scene as a function with a 5D vector of inputs, including the 3D coordinate position $\mathbf{x} = (x, y, z)$ of a spatial point, and the corresponding camera view direction $\mathbf{d} = (\theta, \phi)$. This neural network is:

$$F_\theta: (\mathbf{x}, \mathbf{d}) \rightarrow (c, \sigma). \quad (1)$$

Subsequently, NeRF outputs the volume density $\sigma$ of the 3D position and the RGB color $c$ of that point associated with the corresponding camera viewing direction. Specifically, $x$ is first fed into the Multi-Layer Perceptions (MLP) network and outputs $\sigma$ along with intermediate feature vectors. The intermediate features and $d$ are then imported into an additional fully connected layer and the colors are predicted. Volume rendering based on the composited values calculated through equation (2) obtained from NeRF is further carried out along multiple camera arrays to render a 2D image at an arbitrary viewpoint.

$$C(\mathbf{r}) = \int_{t_n}^{t_f} T(t)\sigma(\mathbf{r}(t))c(\mathbf{r}(t), \mathbf{d})\, dt, \text{where } T(t) = exp\left(-\int_{t_n}^{t_f} \sigma(\mathbf{r}(s))ds\right). \quad (2)$$

However, the volumetric representation of NeRF is based on accumulated volume density $\sigma$ and is unable to reconstruct high-quality surfaces of the object, leading to falsified results when extracting the scene meshes[4].

To address the above problem, two ways are used to reconstruct high-quality 3D surfaces, including the introduction of extra information into the traditional NeRF system and combining different implicit fields. One is to reconstruct high-quality 3D surfaces is to using images and camera poses as supervision to train the implicit fields[3]. These methods indirectly optimize the difference between the shape of a reconstructed scene and the real scene by propagating the evaluated color loss. To further improve the quality of reconstructed scenes, some works[4,5] introduced additional data for supervision based on the traditional mode of image and camera pose supervision. D. Azinović et al.[4] use depth images taken by consumer-level depth cameras. Moreover, DS-NeRF[5] utilizes the generated sparse 3D point cloud by traditional structure-from-motion[6] tools to help solve incorrect geometries. Another approach is to combine other implicit fields with radiance fields, such as Signed Distance Field (SDF)[3] or Truncated Signed Distance Field (TSDF)[4], to jointly express the surface information of a 3D scene, where the 3D geometry can be extracted as a zero-level set of SDF representation. Thus, the volume density in NeRF will be replaced by the (T)SDF value, and the weights in equation (2) will be re-expressed as a function of the (T)SDF value.

Despite the promising results of the SDF methods on top of NeRF, using only the distance fields expressed as functions of three-dimensional coordinates to optimize neural models will encounter serious problems in carving out clear surface boundaries due to the ambiguity introduced in distance field training.

We propose a new method, OmniNeRF, in this paper, which constructs hybrid implicit fields to provide more accurate reconstruction on the edges of the object surface. Specifically, we first address the systematic error in the scene represented by the distance field and the additional bias introduced by the SDF optimization process. OmniNeRF then uses a new form of omnidirectional distance field[7] to represent the geometry, which naturally eliminates errors. Our method, therefore, reconstructs high-quality surfaces, especially around the edges. In summary, our contributions are:

- Addressing a common problem in existing 3D scene reconstruction methods based on hybrid implicit field representations.
- Proposing a novel 3D representation method, OmniNeRF, outputting the omnidirectional distance to the surface for any ray pointing from any direction.

## RELATED WORK

Our method represents the density distribution of a scene by utilizing an omnidirectional distance field. It is related to a novel neural surface reconstruction method based on NeuS[3] and 3D reconstruction methods[6,8].

### 1.1. 3D Reconstruction by Hybrid Implicit Fields

3D reconstruction can be categorized as active-based methods or passive-based methods, Moiré fringe method[8] uses the interference image generated by the overlapping of fence-like fringes. Passive methods in 3D reconstruction obtain images by generating the surrounding environment and inferring the 3D structure of objects. Wang et al.[3] proposed a neural surface reconstruction approach who solves the problem that existing methods tend to fall into local minima. NeuS can control the NeRF volume rendering close to the surface area with its carefully devised weight function that can reconstruct both unbiased and occlusion-aware surfaces. As a result, using the implicit surface method greatly optimizes the performance of the NeRF model on view synthesis. D. Azinović et al.[4] used truncated signed distance fields (TSDF) to represent the scene geometry and used a volumetric integration method to convert TSDF value with the NeRF rendering framework. However, these methods reconstruct blurred edges of a surface.

### 1.2. 3D Reconstruction by Voxels, Point Clouds, and Meshes

Before modern 3D reconstruction methods based on implicit representation, classical 3D reconstruction approaches can generally be divided into three classes according to the data structure of the reconstructed 3D geometry: point cloud-based, voxel-based, and mesh-based methods. Point clouds are a widely used data structure for representing 3D geometry. Previous works[10] based on point clouds 3D reconstruction methods featured by PointNet[10] successfully apply a neural network to address varied 3D recognition tasks suchlike object classification. However, point cloud-based learning methods could not well describe the topology and therefore are difficult to produce watertight surfaces[2]. In contrast to point clouds, voxels can represent any 3D topology, but voxel-based methods require very high computational and storage capacities, which prevents them from retaining fine-grained shape details[11]. Mesh-based methods have shown very promising reconstruction performance[12], but they require predefined template topologies and are therefore challenging to reconstruct shapes with varying topologies.

# METHODS

In this section, we first demonstrate two systematic biases that we observed in current neural representations in 3D reconstruction (Sec. 3.1 and Sec. 3.2). In the second half, we describe a new method, OmniNeRF, to solve this problem (Sec. 3.3).

**1.3. Systematic Biases in SDF-based Scene Representation**

An SDF function evaluates the distance between a point in 3D space and its closest surface, where the SDF value outside of a surface is always positive and the points inside a surface have negative SDF values. When SDF is combined with neural radiance fields, the accumulation formula is re-defined as:

$$C(\mathbf{r}) = \int_{t_n}^{t_f} w\left(f_{SDF}(\mathbf{r}(t))\right) c(\mathbf{r}(t)) dt, \tag{3}$$

where $f_{SDF}$ is the neural implicit signed distance function and $w$ maps signed distance values to the weight in the accumulation. The bell-shaped probabilistic distribution functions are always selected for $w$, and close-to-zero values are mapped to higher weights and vice versa.

With the general framework of combined signed distance and radiance fields, we observed a systematic error in rendering ray colors, as shown in figure 1. When calculating weights on rays close to surface edges, the rendered colors will be blurred if the SDF values are used to calculate the weights because the distance from the surface to the point is much larger than the SDF values. Therefore, inappropriately small weights will be generated when computing the colors cumulatively.

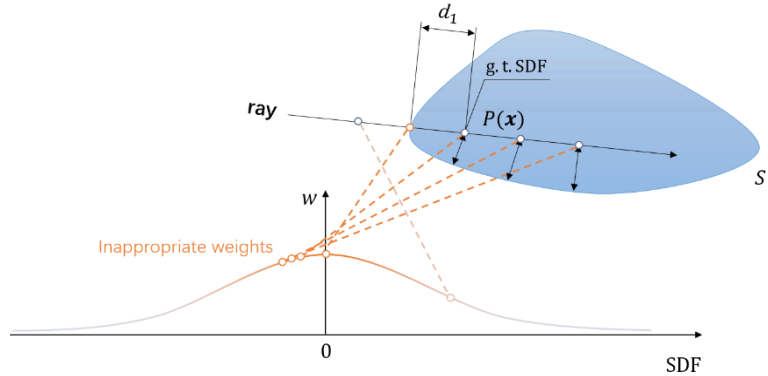

Figure 1. A visualized example for the SDF error. S is a 2D surface; the incoming ray, ideally, accumulates the sampled point color to render the ray color with the weight calculated from the signed distance value along the ray. However, when the point is close enough to the surface, but the ray being queried comes from another direction in 3D space and is far away from the given point (d>>SDF), the weight calculated based on the SDF value is much larger than the weight the point should take in the rendering process.

**1.4. Extra Biases Introduced by SDF Optimization**

Due to the high computational complexity to calculate SDF value in training time, some real-world implementations of hybrid distance and radiance fields (e.g., the implementation codebase of Neural RGB-D Surface Reconstruction[4]) use simplified SDF values to train the implicit signed distance function. A common approach to optimizing the neural signed distance function is estimating the ground-truth SDF value with the distance from the surface to the sampling point and then evaluate the error for backpropagation with L2 loss, which greatly reduces the complexity of model optimization. In this scenario, as illustrated in figure 2, we observed a systematic error between the trained signed distance function and the real-world 3D geometry, i.e., when multiple camera rays intersect at the same point inside of a 3D surface, different signed distance estimates that may differ significantly in magnitude are all taken to optimize the scalar SDF values of the same point. As a result, the SDF representation will not easily converge to an appropriate distribution, which in turn will further degrade the reconstruction quality.

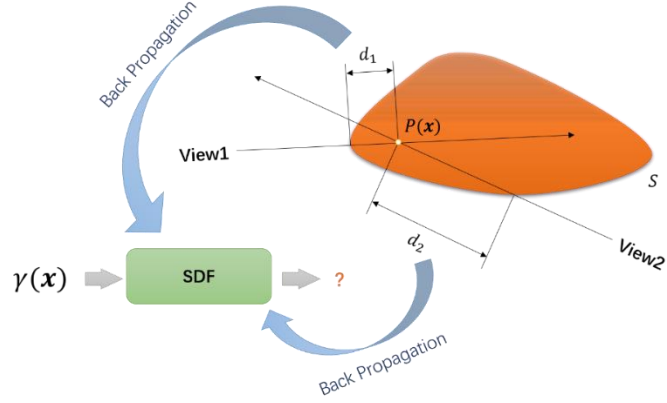

Figure 2. A visualized example of the systematic ambiguity between the simplified calculations of signed distance values. S is a 2D surface (just for visualization); View1 and View2 are two different rays intersecting at the 2D-point P(x). In common SDF implementations, the distance between the ground-truth surface and the querying point depth (i.e., $d_1$, $d_2$ in the figure) is considered as "ground truth SDF value", and although neither of them is the real signed distance value of point $P(x)$, they are both used in the backpropagation process to train the neural SDF representation.

### 1.5. OmniNeRF

In this work, we proposed OminiNeRF for 3D surface reconstruction, which formulates as an optimization process. Given a set of posed images $\{I_i\}$, $\{R_i, t_i\}$ and the aligned depth maps$\{D_i\}$, our goal is to reconstruct high-fidelity surface $S$ of them. The depth images may be obtained from consumer-level depth cameras. We will use the N color image frames along with the aligned depth images and ground-truth camera poses captured at the same time.

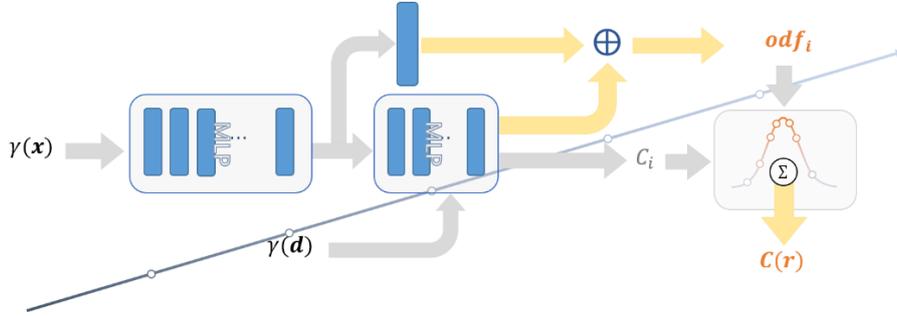

Figure 3. The architecture of the OmniNeRF model.

In OmniNeRF, as illustrated in figure 3, we represent the scene geometry as an omnidirectional distance field (ODF), i.e., a 5D function:

$$d = f(\mathbf{x}, \mathbf{d}) = f_{SDF}(x,y,z) + f_{ODF}(x,y,z,\theta,\phi). \tag{4}$$

The embedded positions are the first input into the first MLP producing intermediate features for the subsequent operations. The omnidirectional distance value is then computed in the fully connected layer and the second MLP (in the middle, which accepts the embedded directions and intermediate features as inputs). Finally, the output colors and ODF values are then combined to calculate the rendering color for the ray.

Specifically, we implement this hybrid scene representation using an MLP which takes embedded coordinates[1] and orientations as input and outputs the omnidirectional distance values as well as the accumulated colors.

The ODF value is represented as a summation of a signed distance value and a correction bias term, in which the SDF values are evaluated only with a 3-dimensional position vector, and the correction terms are produced using the full input camera poses. When querying a ray color, the weights are computed with the ODF values[4]:

$$w = w\big(f(\mathbf{x},\mathbf{d})\big) = \sigma\left(\frac{f(\mathbf{x},\mathbf{d})}{\text{tr}}\right)\sigma\left(-\frac{f(\mathbf{x},\mathbf{d})}{\text{tr}}\right), \tag{5}$$

where $\sigma$ is the sigmoid function to convert the distance value into the $\alpha$-compositional weight and tr is the truncation distance like TSDF representations.

A positional encoding[13] is applied before passing camera poses into the neural network to help our implicit representation better learn the high-frequency details of the scene:

$$\gamma(x) = (x, \sin(2^0 x), \cos(2^0 x), \ldots, \sin(2^k x), \cos(2^k x)). \tag{6}$$

**1.6. Optimization**

To optimize our network, we randomly sample a batch of $P_c$ points along projective rays generated based on input camera poses. An extra batch of $P_f$ sampling points based on the coarse observation of the distribution of the ODF values evaluated on the first $P_c$ points.

$$L(P) = \lambda_1 L_C(P) + \lambda_2 L_D(P) + \lambda_3 L_{ODF}(P). \tag{7}$$

**Color Loss:** The color loss $L_C(P)$ evaluates the MSE value between the rendered and the inputted color, which can be defined as:

$$L_C(P) = \mathrm{MSE}(f_\Theta(\gamma(\mathbf{x}), \gamma(\mathbf{d})), \hat{C}), \tag{8}$$

where $C = f_\Theta(\gamma(\mathbf{x}), \gamma(\mathbf{d}))$ is the accumulated rendered color and $\hat{C}$ is the observed color.

**Depth Loss:** The depth loss in (8) measures the error of the predicted depth image, where the rendered depth is calculated through (9), which helps the distance fields to converge.

$$L_D(P) = \mathrm{MSE}(D, \widehat{D}). \tag{9}$$

$$D(\mathbf{x}, \mathbf{d}) = \int_{t_n}^{t_f} t \cdot w(\gamma(\mathbf{x}), \gamma(\mathbf{d})) dt. \tag{10}$$

**ODF Loss:** The ODF Loss can be further separated into three components, in which $L_{fs}$ and $L_{tr}$ help regularize the SDF submodule and the $L_{ODF}$ term force the ODF network to learn predict the correct $f'_{ODF}(\mathbf{x}, \mathbf{d})$ bias term.

$$f_{ODF}(\mathbf{x}, \mathbf{d}) = f_{TSDF}(\mathbf{x}) + f'_{ODF}(\mathbf{x}, \mathbf{d}), \tag{11}$$

$$f_{TSDF}(\mathbf{x}) = \min_{\mathbf{d}}(f_{ODF}(\mathbf{x}, \mathbf{d})). \tag{12}$$

Because $f_{TSDF}$ is a piecewise function, we need to supervise the parts of the distance inside and outside the $[-tr, tr]$ range near the surface.

$$L_{fs}(P) = \frac{1}{|P_{fs}|} \Sigma_{p_{fs}} (f_{TSDF}(\mathbf{x}), tr). \tag{13}$$

$$L_{tr}(P) = \frac{1}{|P_{tr}|} \Sigma_{p_{tr}} (f_{TSDF}(\mathbf{x}) - \hat{d}). \tag{14}$$

## EXPERIMENT SETTINGS

This section introduces the dataset, baseline methods, and metrics we used to evaluate performances.

**1.7. Dataset**

Our method is evaluated on the ScanNet dataset, same as Neural RGB-D[4]. The ScanNet dataset contains various types of indoor scenes, which includes different geometries, and is usually used to evaluate indoor 3D reconstruction methods. A series of colored images, the aligned depth maps, and the corresponding ground-truth camera trajectories are used to optimize the model.

**1.8. Baseline Methods**

We examine our approach with cutting-edge methods: 1. Neural Radiance Fields (NeRF), which adopt neural fields to render photo-realistic novel views; 2. Neural RGB-D Surface Reconstruction[4], which uses depth supervision to help reconstruct 3D scene geometry.

### 1.9. Metrics

Our model performance is measured from both image representation and 3D structure reconstruction. The image representation quantitatively evaluates image outputs on a specific camera pose and position, while the 3D structural reconstruction intuitively reveals the overall performance of our surface reconstruction. We applied PSNR values to calculate the difference between the outputs and observed images. To evaluate our reconstructed surface and models, several metrics including RMSE, Abs Rel, Sq Rel, $\delta_1, \delta_2,$ and $\delta_3$ are measured with the following:

$$\text{RMSE} = \sqrt{\frac{1}{N}\sum_{d^*}\|d^* - d\|^2}, \tag{15}$$

$$\delta_i = \frac{1}{N}|\{d^*| \max\left(\frac{d^*}{d}, \frac{d}{d^*}\right) < 1.25^i\}|, \tag{16}$$

where $d$ and $d^*$ represent the accumulated and the inputted depth values and D is the set of inputted depth maps.

## RESULTS AND ANALYSIS

We evaluated our models by visual comparison between the image outputs between the reconstructed and the observed, a quantitative comparison based on the metrics in section 1.9. , and an ablation study where various network architectures to implement ODF representation are tested.

### 1.10. Visual Evaluation

We first evaluated our models by comparing the difference among image outputs.

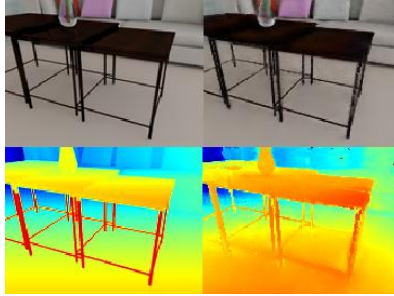

Figure 4. Visual output of our model Omni-NeRF (Top-left: ground-truth RGB image; Top-right: Our rendered image; Bottom-left: input depth map; Bottom-right: predicted depth map)

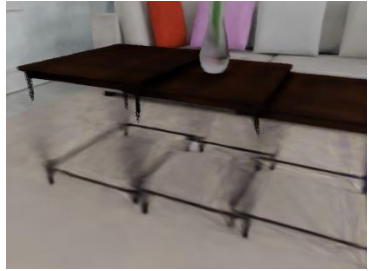

Figure 5. Failure case of previous surface reconstruction method[4]

### 1.11. Quantitative Evaluation

The benchmark evaluations can be seen in Tab. 1, where OmniNeRF obtains better results in different testcases.

Table 1. Reconstruction results of baseline methods and our model on ScanNet[9].

| Method | PSNR↑ | RMSE | $\delta_1$ | $\delta_2$ | $\delta_3$ |
|---|---|---|---|---|---|
| NeRF | 26.56 | 0.087 | 0.83 | 0.905 | 0.91 |
| Neural RGB-D | 26.87 | 0.130 | 0.878 | 0.879 | 0.881 |
| **Ours** | **29.01** | **0.077** | **0.882** | **0.929** | **0.939** |

**1.12. Ablation Study**

We tested various network architectures to implement ODF representation: (a) As an output of the rendering network. (b) Predicted by using intermediate features of NeRF as well as viewing directions.

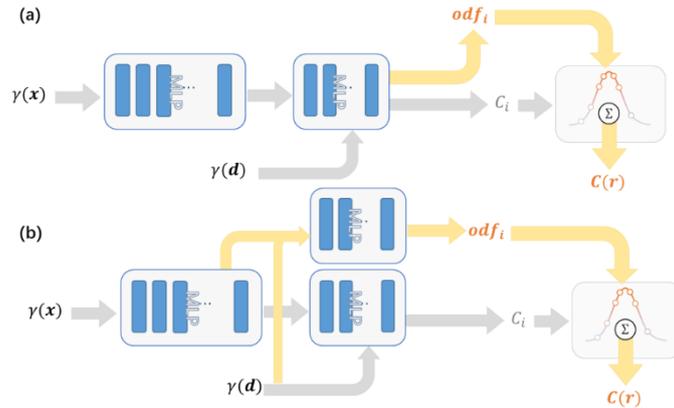

Figure 6. Alternative network structures.

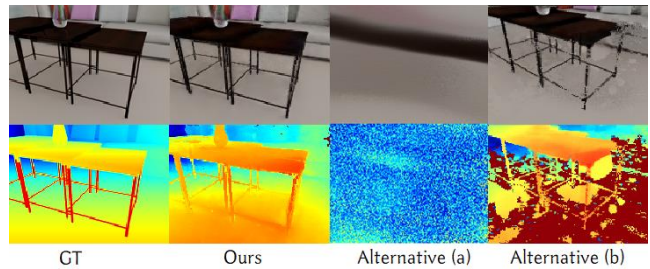

Figure 7. The visualization result on different network structures.

## CONCLUSIONS

We presented OmniNeRF for 3D surface reconstruction by introducing a hybrid implicit field including a neural radiation field (NeRF) and an omnidirectional distance field (ODF). We use an MLP architecture to model the scene geometry using a set of RGB-D images and achieve cutting-edge performance compared to baseline models.

By effectively producing high-quality 3D surfaces, our hybrid implicit 3D reconstruction method can provide more realistic reconstruction results, thus enabling more accurate detection or identification of objects. As a result, our method may have very promising applications and impacts in automotive driving, virtual reality, and augmented reality.

Our method achieves both surface reconstruction with high fidelity and photo-realistic novel view synthesis. Our analysis demonstrates that the transformation from signed distance field representations to omnidirectional distance fields improves the reconstruction quality at the edges of surfaces. One limitation of our method, like other NeRF-based methods, is that the time costs of training and inferencing the model may take a long time. However, according to recent research, this problem can be addressed by adopting explicit representations such as a multi-resolution hash table[14]. Combining our method with these methods may be one future topic.